\documentclass[sigconf]{acmart}
\AtBeginDocument{%
  \providecommand\BibTeX{{%
    \normalfont B\kern-0.5em{\scshape i\kern-0.25em b}\kern-0.8em\TeX}}}

\usepackage{multirow}
\usepackage{amsmath}
\usepackage{algorithm}
\usepackage{algorithmic}

\copyrightyear{2022} 
\acmYear{2022} 
\setcopyright{acmcopyright}\acmConference[MM '22]{Proceedings of the 30th ACM International Conference on Multimedia}{October 10--14, 2022}{Lisboa, Portugal}
\acmBooktitle{Proceedings of the 30th ACM International Conference on Multimedia (MM '22), October 10--14, 2022, Lisboa, Portugal}
\acmPrice{15.00}
\acmDOI{10.1145/3503161.3548547}
\acmISBN{978-1-4503-9203-7/22/10}

\begin{document}

\title{DavarOCR: A Toolbox for OCR and Multi-Modal Document Understanding}

\author{Liang Qiao}
\affiliation{%
  \institution{Hikvision Research Institute}
  \city{Hangzhou}
  \country{China}
}

\author{Hui Jiang}
\affiliation{%
  \institution{Hikvision Research Institute}
  \city{Hangzhou}
  \country{China}
}

\author{Ying Chen}
\affiliation{%
  \institution{Hikvision Research Institute}
  \city{Hangzhou}
  \country{China}
}

\author{Can Li}
\affiliation{%
  \institution{Hikvision Research Institute}
  \city{Hangzhou}
  \country{China}
}
\author{Pengfei Li}
\affiliation{%
  \institution{Hikvision Research Institute}
  \city{Hangzhou}
  \country{China}
}
\author{Zaisheng Li}
\affiliation{%
  \institution{Hikvision Research Institute}
  \city{Hangzhou}
  \country{China}
}
\author{Baorui Zou}
\affiliation{%
  \institution{Hikvision Research Institute}
  \city{Hangzhou}
  \country{China}
}
\author{Dashan Guo}
\affiliation{%
  \institution{Hikvision Research Institute}
  \city{Hangzhou}
  \country{China}
}
\author{Yingda Xu}
\affiliation{%
  \institution{Hikvision Research Institute}
  \city{Hangzhou}
  \country{China}
}
\author{Yunlu Xu}
\affiliation{%
  \institution{Hikvision Research Institute}
  \city{Hangzhou}
  \country{China}
}
\author{Zhanzhan Cheng$^*$} 
\affiliation{%
  \institution{Hikvision Research Institute}
  \city{Hangzhou}
  \country{China}
}
\author{Yi Niu}
\affiliation{%
  \institution{Hikvision Research Institute}
  \city{Hangzhou}
  \country{China}
}

\renewcommand{\shortauthors}{Qiao, et al.}

\begin{abstract}
  This paper presents DavarOCR, an open-source toolbox for OCR and document understanding tasks. DavarOCR currently implements 19 advanced algorithms, covering 9 different task forms. DavarOCR provides detailed usage instructions and the trained models for each algorithm. Compared with the previous open-source OCR toolbox, DavarOCR has relatively more complete support for the sub-tasks of the cutting-edge technology of document understanding. In order to promote the development and application of OCR technology in academia and industry, we pay more attention to the use of modules that different sub-domains of technology can share. DavarOCR is publicly released at \url{https://github.com/hikopensource/Davar-Lab-OCR}.
\end{abstract}




\maketitle
\newcommand{\tabincell}[2]{\begin{tabular}{@{}#1@{}}#2\end{tabular}}
\section{Introduction}
OCR (Optical Character Recognition) is an essential technique in deep learning and has been applied in many fields like finance, transportation, and health care. With the continuous development of technology, people's demand for OCR ability has changed from simple text content extraction to more diversified and intelligent tasks that can directly solve the problems in actual production. Based on this, we divide the OCR tasks into two types, the \emph{basic OCR} tasks and \emph{document understanding} tasks. 

The basic OCR tasks are the \emph{perception} tasks that aim to obtain the content of the text in the images/videos and convert it into the electronic format, usually including text detection \cite{zhou2017east,baek2019character,DBLP:conf/aaai/LiaoWYCB20} and text recognition \cite{CRNN,DBLP:conf/cvpr/FangXWM021,DBLP:conf/kdd/BorisyukGS18}. To be more efficient and robust, people also propose end-to-end architectures \cite{li2017towards,liao2019mask,DBLP:conf/aaai/QiaoTCXNPW20}. However, solely capturing text content 
sometimes cannot directly address the actual needs. 
In actual OCR production, intelligent systems need to perform a variety of downstream tasks based on the results of OCR. We call these tasks document understanding. These tasks can be regarded as \emph{cognitive} tasks that require people to further understand the information (semantic information or relational information) conveyed by the text of the document. For example, Key Information Extraction (KIE) \cite{DBLP:conf/emnlp/KattiRGBBHF18,DBLP:conf/mm/ZhangXCP0QNW20,DBLP:journals/corr/abs-2103-14470} is one of the common tasks in the financial instrument identification system, where the model needs to identify the values of the specific items like `invoice code' and `amount'.  Other common forms of document understanding tasks include layout analysis \cite{DBLP:conf/cvpr/YangYAKKG17,DBLP:conf/icdar/ZhangLQCPN021}, Reading Order Detection (ROD) \cite{DBLP:conf/eccv/LiGBWYZ20,wang2021layoutreader}, table recognition \cite{khan2019table,DBLP:conf/icdar/QiaoLCZPNRT021} and understanding \cite{wang2020structure,DBLP:conf/acl/JauharTH16}, document Question-Answering (QA) \cite{DBLP:conf/wacv/MathewKJ21}, Named Entity Recognition (NER) \cite{DBLP:conf/conll/SangM03,DBLP:conf/acl/YanGDGZQ20,DBLP:conf/sigir/GuoXCL09,DBLP:journals/corr/HuangXY15}, etc.

\begin{table*}[t]

    \caption{Comparison of support algorithms between different open-source OCR toolboxes. All methods in Basic OCR Tasks only use visual information. In Document Understanding tasks, methods labelled with $\heartsuit$ means using \emph{visual} features, with $\clubsuit$ means using \emph{textual} features, with $\diamondsuit$ means using \emph{positional} features. }
    \label{tab:ablation_components}
    \scalebox{0.8}{
    \begin{tabular}{c|cccc|ccccc}
      \toprule
    \multirow{3}{*}{Toolbox}  & \multicolumn{4}{c|}{Basic OCR Tasks} & \multicolumn{5}{c}{Document Understanding Tasks} \\
     \cline{2-10}
     & \tabincell{c}{Text \\ Detection} & \tabincell{c}{Text \\ Recognition} & \tabincell{c}{End-to-End \\ Text Spotting} & \tabincell{c}{Video \\ Text} &  \tabincell{c}{KIE} & \tabincell{c}{NER}  & \tabincell{c}{Layout \\ Analysis} & \tabincell{c}{ROD} & \tabincell{c}{Table \\ Recognition} \\
      \midrule
      tesseract & convention &
       \tabincell{c}{LSTM} & - & -
       & - & - & - & - & - \\
        \midrule
         \tabincell{c}{chineseocr} & \tabincell{c}{Yolo\cite{DBLP:conf/cvpr/RedmonF17}} &
       CRNN\cite{CRNN} & - & -
       & - & - & - & - & -  \\
        \midrule
         \tabincell{c}{chineseocr\_lite} & \tabincell{c}{DB\cite{DBLP:conf/aaai/LiaoWYCB20}} &
       CRNN\cite{CRNN} & - & -
       & - & - & - & - & -  \\
       \midrule
        \tabincell{c}{EasyOCR} & \tabincell{c}{CRAFT\cite{baek2019character}} &
       CRNN\cite{CRNN} & - & -
       & - & - & - & - & -  \\
       \midrule
       PaddleOCR & 
       \tabincell{c}{DB\cite{DBLP:conf/aaai/LiaoWYCB20}, \\ EAST\cite{zhou2017east}, \\ etc. (4)} &
       \tabincell{c}{CRNN\cite{CRNN}, \\ Rosetta\cite{DBLP:conf/kdd/BorisyukGS18} \\ etc. (7) } & 
       \tabincell{c}{PGNet\cite{DBLP:conf/aaai/WangZQLZLHLDS21}} & -
       & \tabincell{c}{SDMGR\cite{DBLP:journals/corr/abs-2103-14470} \\ $\heartsuit\clubsuit\diamondsuit$} & -  & Yolo\cite{DBLP:conf/cvpr/RedmonF17}$\heartsuit$ & - & RARE\cite{DBLP:conf/cvpr/ShiWLYB16}$\heartsuit$  \\
        \midrule
       mmocr & \tabincell{c}{DB\cite{DBLP:conf/aaai/LiaoWYCB20}, \\ M-RCNN\cite{he2017mask}, \\ etc. (6)} &
       \tabincell{c}{CRNN\cite{CRNN}, \\ ABINet\cite{DBLP:conf/cvpr/FangXWM021} \\ etc. (8) } & - & - & \tabincell{c}{SDMGR\cite{DBLP:journals/corr/abs-2103-14470} \\ $\heartsuit\clubsuit\diamondsuit$} & \tabincell{c}{BERT - \\ Softmax\cite{devlin2019bert}$\clubsuit$} & - & - & -  \\
       \midrule
       davarocr & \tabincell{c}{EAST\cite{zhou2017east}, \\ M-RCNN\cite{he2017mask}, \\ TP-det\cite{DBLP:conf/aaai/QiaoTCXNPW20}}
       & \tabincell{c}{CRNN\cite{CRNN}, \\ SPIN\cite{DBLP:conf/aaai/0003XCPNWZ21},\\etc. (5)} & \tabincell{c}{TP\cite{DBLP:conf/aaai/QiaoTCXNPW20}, \\ MANGO\cite{DBLP:conf/aaai/QiaoCCXNPW21}, \\ M-RCNN-e2e} & YORO \cite{DBLP:conf/mm/ChengLNP0Z19} & \tabincell{c}{Chargrid\cite{DBLP:conf/emnlp/KattiRGBBHF18}$\heartsuit\clubsuit$, \\  TRIE\cite{DBLP:conf/mm/ZhangXCP0QNW20}$\heartsuit\clubsuit\diamondsuit$} & 
       \tabincell{c}{ BERT-Softmax \\ /Span/CRF\cite{devlin2019bert} $\clubsuit$ \\ BiLSTM-CRF\cite{DBLP:journals/corr/HuangXY15}$\clubsuit$} & VSR\cite{DBLP:conf/icdar/ZhangLQCPN021}$\heartsuit\clubsuit\diamondsuit$ & \tabincell{c}{GCN-PN\cite{DBLP:conf/eccv/LiGBWYZ20}$\heartsuit\diamondsuit$} & \tabincell{c}{LGPMA\cite{DBLP:conf/icdar/QiaoLCZPNRT021}$\heartsuit$} \\ 
    \bottomrule
    \end{tabular}
    }
\end{table*}

Nowadays, many advanced open-source OCR toolboxes aim to provide more efficient and unified OCR services. Tesseract\footnote{https://github.com/tesseract-ocr/tesseract}, chineseOCR\footnote{https://github.com/chineseocr/chineseocr}, chineseOCR\_lite\footnote{https://github.com/DayBreak-u/chineseocr\_lite}, EasyOCR\footnote{https://github.com/JaidedAI/EasyOCR} are some earlier OCR toolboxes. They are mostly focused on providing an efficient OCR inference service and only support the tasks of text detection and recognition. PaddleOCR\footnote{https://github.com/PaddlePaddle/PaddleOCR} is a popular open-source OCR toolbox based on PaddlePaddle, which also provides relatively complete services for OCR production like labeling, training, inference and deployment. In some of its latest versions (up to v2.2), PaddleOCR starts to provide the support of some document understanding tasks, including KIE, Layout Analysis, Table Recognition and Document VQA. In each task at this stage, they provide some fairly basic and simple implementations, \emph{e.g.}, a basic object detection solution (Yolo\cite{DBLP:conf/cvpr/RedmonF17}) for layout analysis. Mmocr\footnote{https://github.com/open-mmlab/mmocr} is a recent Pytorch-based OCR toolbox that collects many implementations of advanced OCR algorithms. It provides a rich and flexible component configuration to build an OCR model, including different backbones, necks, heads, losses, etc. Currently, mmocr has little support for document understanding tasks, \emph{i.e.}, an KIE (SDMGR\cite{DBLP:journals/corr/abs-2103-14470}) model and NER (BERT-Softmax\cite{devlin2019bert}) model. 

In this paper, we present DavarOCR, a new OCR toolbox that can well support the vast majority of current basic OCR and document understanding tasks. Similar to mmocr, DavarOCR is also built based on the training engine mmcv\footnote{https://github.com/open-mmlab/mmcv} and the architecture design of mmdetection\footnote{https://github.com/open-mmlab/mmdetection}. This means the whole framework can be compatible with mmocr. Therefore, we do not implement many methods for text detection and recognition algorithms but focus on expanding the tasks that MMOCR cannot support. For the basic OCR tasks, besides other text detection and recognition algorithms, we extend the framework to well support end-to-end text spotting and video text tasks. Notably, DavarOCR contains complete modules for end-to-end text spotting tasks, including both two-staged  (TP\cite{DBLP:conf/aaai/QiaoTCXNPW20},Mask-RCNN-based) and one-staged (MANGO\cite{DBLP:conf/aaai/QiaoCCXNPW21}) text spotting architectures. For the document understanding tasks, we extend the framework's support for some new tasks, including Layout Analysis, ROD, and Table Recognition. Shortly, we will continue to expand the implementation of new tasks, including Document VQA and Table Understanding. At the present stage, DavarOCR contains the implementations of 19 advanced algorithms, including 10 algorithms that any other previous framework has not implemented.  An algorithm-level comparison between the open-source toolboxes is shown in Table \ref{tab:ablation_components}.

DavarOCR is released under the Apache-2.0 License. The code repository contains the one-click setup script, accompanied by the complete demo and detailed instruments to make researchers understanding easier.

\begin{figure*}[t]
  \centering
  \includegraphics[width=1\linewidth]{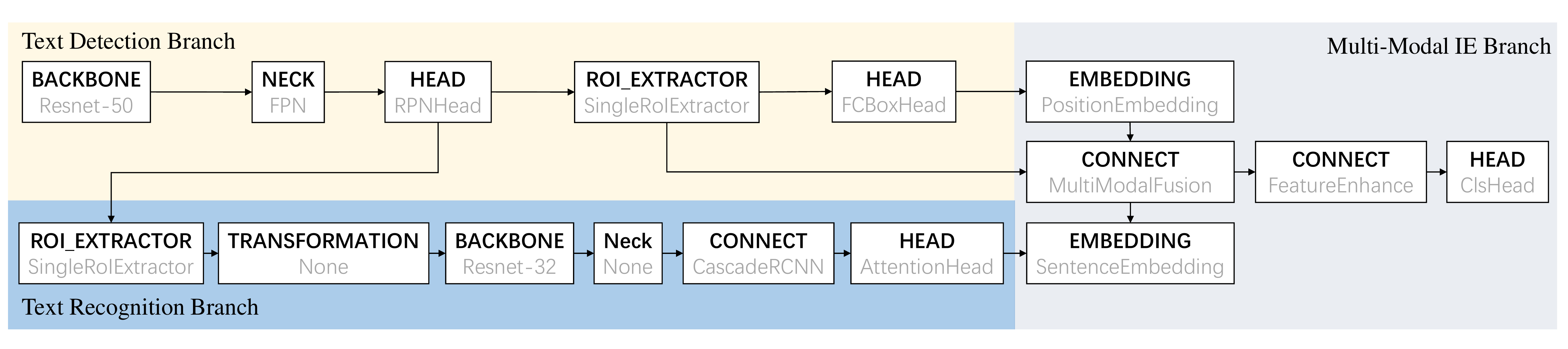}
  \caption{The modular design for the end-to-end KIE model TRIE.}
  \label{fig:trie}
\end{figure*}

\section{Highlight Features of DavarOCR}
\subsection{Across Tasks Module Sharing}
DavarOCR inherits the idea of modular design from mmdetection \cite{DBLP:journals/corr/abs-1906-07155} and extends it to support more tasks. In mmdetection, a model can be divided into 4 four parts: BACKBONE, NECK, ROI\_EXTRACTOR and HEAD. We can modify the configuration to make the model be integrated with an arbitrary combination of modules, such as different backbones. Although text detection models can mostly follow the same module divisions, other tasks will inevitably introduce new modules. In DavarOCR, some of the extended model-related modules include:

\emph{TRANSFORMATION}. In text recognition, some algorithms will integrate a learnable deformation correction module to rectify the text with irregular shapes, such as Affine\cite{DBLP:conf/nips/JaderbergSZK15}, TPS\cite{DBLP:conf/cvpr/ShiWLYB16}. These modules can also be used in text spotting, video text tasks.

\emph{EMBEDDING}. In some tasks that require textual or positional information, the model needs to utilize the features encoded by an embedding layer. For example, word/sentence embedding is used in representing the textual feature, and 2D positional embedding is used to establish spatial location connections. The related tasks include KIE, NER, Layout Analysis, ROD, etc.

\emph{CONNECT}. Some intermediate connection modules are used frequently in a model like feature enhancement, feature deformation, and feature fusion. For example, Bi-LSTM\cite{hochreiter1997long} is a typical module used to enhance sequence features, which can be used in any text-recognition-related (text spotting, KIE, etc.) and NER tasks.

 In addition to the above mentioned model-related components, many commonly used modules/operators/tools can be found in DavarOCR. All of the above modular design in DavarOCR allows for maximum component sharing between different tasks, enabling researchers to build a cross-task model quickly. Figure \ref{fig:trie} demonstrates an example architecture of the modal TRIE\cite{DBLP:conf/mm/ZhangXCP0QNW20}, which is a end-to-end model contains three sub-tasks: text detection, recognition and information extraction.

\subsection{Uniform Data Label Format}
Because of the different types of tasks, we often need to use different supervision to train models, which leads to varying forms of task-dependent data annotation. In fact, there are many similarities between basic OCR and document understanding tasks. To further unify the data processing logic in DavarOCR, We try to integrate all possible tasks and propose some unified basic formats for the data label.

\begin{figure}[ht]
  \centering
  \includegraphics[width=0.8\linewidth]{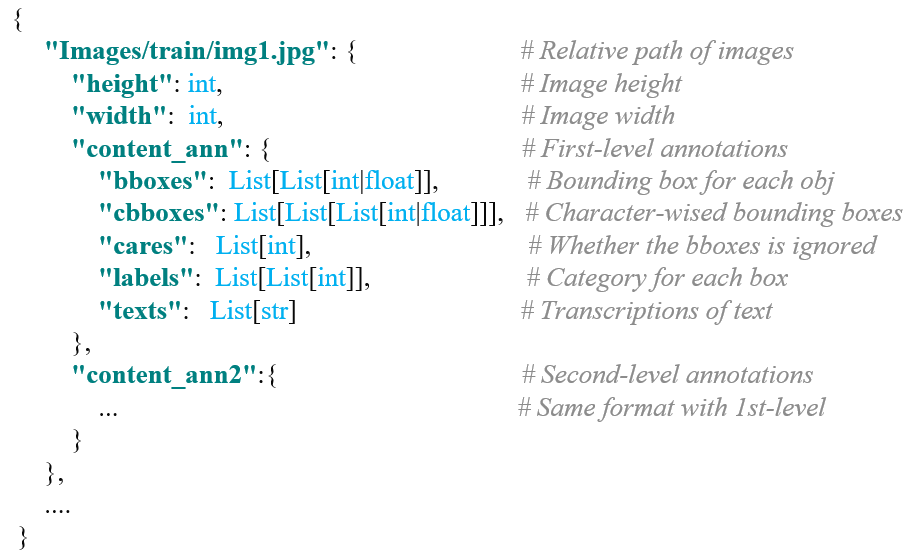}
  \caption{Illustration of the basic data label}
  \label{fig:ann}
  \Description{Illustration of the basic image-based data label format.}
\end{figure}

Figure \ref{fig:ann} shows an illustration of the basic image-based data label format. In the basic format, annotation information for all images is arranged in a JSON file, and the keys are the images' paths. The detailed annotation is stored in the item of ``content\_ann''. All the items in ``content\_ann" are array types and have the same length on the outermost level. 
This indicates that all the labeling information corresponds to each bounding box. If someone needs to store full-image-level annotations, all the lists are of length 1, and ``bboxes'' default to contain a single box to represent the entire image range. 
Item of ``labels'' saves the category information for an object or the whole image, which is used in the tasks related to object classification like KIE and Layout Analysis. 
It is stored in a two-dimensional list to support the classification of multitasking.
Notably, we add keywords (``content\_ann2'') to save multilevel labels for supporting the multilevel task.
For example, in the multi-modal layout analysis task, the model needs to simultaneously learn the page-level (block-level) and text-line-level tasks.

The basic data label format supports most image-based tasks, including text detection, text recognition, end-to-end text spotting, KIE, and Layout Analysis. When we use the data to train a model, we can freely select any combination of the keywords to form the required information for the task. We can define the task-specific data form for new tasks that inherit the basic format to supplement the required items. For example, in table recognition tasks, the model relies on the item to represent cell indexes. DavarOCR also provides the data format for the video-based (Video Text) of plain-text-based (NER) tasks.

\begin{table}[ht]
    \caption{Multi-modal ablation results of the Key Information Extraction task on WildReceipt\cite{DBLP:journals/corr/abs-2103-14470}.}
    \label{tab:ablation1}
    \scalebox{0.85}{
    \begin{tabular}{c|c|c|c|c}
      \toprule
    Algorithm  & Modalities & Backbone & \tabincell{c}{Testing\\ Scale} & F1-Score   \\
    \midrule
    \multirow{2}{*}{Chargrid\cite{DBLP:conf/emnlp/KattiRGBBHF18}} & Visual & ChargridNet & (512,512) & 67.10 \\
    & \tabincell{c}{Visual+Textual} & ChargridNet & (512,512) & 81.02 \\ 
    \midrule
    \multirow{5}{*}{TRIE\cite{DBLP:conf/mm/ZhangXCP0QNW20}} & \tabincell{c}{Visual} & ResNet-50 & (512,512) & 78.25 \\ 
     & \tabincell{c}{Visual+Textual} & ResNet-50 & (512,512) & 84.67 \\ 
     & \tabincell{c}{Visual+Positional} & ResNet-50 & (512,512) & 86.57 \\ 
     & \tabincell{c}{Visual+Textual+\\Positional} & ResNet-50 & (512,512) & 87.08 \\ 
    \bottomrule
    \end{tabular}
    }
\end{table}
\begin{table}[ht]
    \caption{Multi-modal ablation results of the Layout Analysis task on PubLayNet\cite{DBLP:conf/icdar/ZhongTJ19}. (C) means Character-level textual feature, and (S) means Sentence-level textual feature. }
    \label{tab:ablation2}
    \scalebox{0.8}{
    \begin{tabular}{c|c|c|c|c}
      \toprule
    Algorithm  & Modalities & Backbone & \tabincell{c}{Testing\\ Scale} & mAP   \\
    \midrule
    \multirow{5}{*}{VSR\cite{DBLP:conf/icdar/ZhangLQCPN021}} & \tabincell{c}{Visual} & ResNeXt-101 & (1300,800) & 92.6 \\ 
     & \tabincell{c}{Visual+Textual (C)} & ResNeXt-101 & (1300,800) & 94.5 \\ 
     & \tabincell{c}{Visual+Textual (S)} & ResNeXt-101 & (1300,800) & 94.7 \\ 
     & \tabincell{c}{Visual+Textual\\ (C+S)+Positional} & ResNeXt-101 & (1300,800) & 95.8 \\ 
    \bottomrule
    \end{tabular}
    }
\end{table}
Through the uniform data format, most of the tasks from different domains can share the same data processing functions like data loading, data augmentation and data formatting.

\section{Case Study: Multi-Modal Document Understanding}

Different modal information may play different roles in document understanding tasks. This section demonstrates some of the modality ablation results in DavarOCR.

Table \ref{tab:ablation1} and Table \ref{tab:ablation2} separately show some models' ablation results on the KIE and Layout Analysis tasks. In the KIE experiments, the task can be simply implemented by a single visual-based model, which equals an object detection task. The Chargrid model uses a char-level segmentation map to introduce textual features, and TRIE will extract the positional and textual features to compromise the box-level multi-modal features. From the results, we can see that involving additional modalities can greatly improve the model's performance.

\section{Conclusion}
This paper presents an open-source toolbox, DavarOCR, for comprehensive OCR and document understanding tasks. DavarOCR currently implements 19 state-of-the-art algorithms, covering 9 different task areas. In DavarOCR, researchers can freely combine the reusable modules and adopt multi-modal information to implement more complex algorithms involving perception and cognition. We hope that this framework can further promote the implementation of new technologies in actual production.

\bibliographystyle{ACM-Reference-Format}
\bibliography{egbib}


\begin{thebibliography}{37}


\ifx \showCODEN    \undefined \def \showCODEN     #1{\unskip}     \fi
\ifx \showDOI      \undefined \def \showDOI       #1{#1}\fi
\ifx \showISBNx    \undefined \def \showISBNx     #1{\unskip}     \fi
\ifx \showISBNxiii \undefined \def \showISBNxiii  #1{\unskip}     \fi
\ifx \showISSN     \undefined \def \showISSN      #1{\unskip}     \fi
\ifx \showLCCN     \undefined \def \showLCCN      #1{\unskip}     \fi
\ifx \shownote     \undefined \def \shownote      #1{#1}          \fi
\ifx \showarticletitle \undefined \def \showarticletitle #1{#1}   \fi
\ifx \showURL      \undefined \def \showURL       {\relax}        \fi
\providecommand\bibfield[2]{#2}
\providecommand\bibinfo[2]{#2}
\providecommand\natexlab[1]{#1}
\providecommand\showeprint[2][]{arXiv:#2}

\bibitem[Baek et~al\mbox{.}(2019)]%
        {baek2019character}
\bibfield{author}{\bibinfo{person}{Youngmin Baek}, \bibinfo{person}{Bado Lee},
  \bibinfo{person}{Dongyoon Han}, \bibinfo{person}{Sangdoo Yun}, {and}
  \bibinfo{person}{Hwalsuk Lee}.} \bibinfo{year}{2019}\natexlab{}.
\newblock \showarticletitle{Character region awareness for text detection}. In
  \bibinfo{booktitle}{\emph{CVPR}}. \bibinfo{pages}{9365--9374}.
\newblock


\bibitem[Borisyuk et~al\mbox{.}(2018)]%
        {DBLP:conf/kdd/BorisyukGS18}
\bibfield{author}{\bibinfo{person}{Fedor Borisyuk}, \bibinfo{person}{Albert
  Gordo}, {and} \bibinfo{person}{Viswanath Sivakumar}.}
  \bibinfo{year}{2018}\natexlab{}.
\newblock \showarticletitle{Rosetta: Large Scale System for Text Detection and
  Recognition in Images}. In \bibinfo{booktitle}{\emph{KDD}}.
  \bibinfo{pages}{71--79}.
\newblock


\bibitem[Chen et~al\mbox{.}(2019)]%
        {DBLP:journals/corr/abs-1906-07155}
\bibfield{author}{\bibinfo{person}{Kai Chen}, \bibinfo{person}{Jiaqi Wang},
  \bibinfo{person}{Jiangmiao Pang}, \bibinfo{person}{Yuhang Cao},
  \bibinfo{person}{Yu Xiong}, \bibinfo{person}{Xiaoxiao Li},
  \bibinfo{person}{Shuyang Sun}, \bibinfo{person}{Wansen Feng},
  \bibinfo{person}{Ziwei Liu}, \bibinfo{person}{Jiarui Xu},
  \bibinfo{person}{Zheng Zhang}, \bibinfo{person}{Dazhi Cheng},
  \bibinfo{person}{Chenchen Zhu}, \bibinfo{person}{Tianheng Cheng},
  \bibinfo{person}{Qijie Zhao}, \bibinfo{person}{Buyu Li}, \bibinfo{person}{Xin
  Lu}, \bibinfo{person}{Rui Zhu}, \bibinfo{person}{Yue Wu},
  \bibinfo{person}{Jifeng Dai}, \bibinfo{person}{Jingdong Wang},
  \bibinfo{person}{Jianping Shi}, \bibinfo{person}{Wanli Ouyang},
  \bibinfo{person}{Chen~Change Loy}, {and} \bibinfo{person}{Dahua Lin}.}
  \bibinfo{year}{2019}\natexlab{}.
\newblock \showarticletitle{MMDetection: Open MMLab Detection Toolbox and
  Benchmark}.
\newblock \bibinfo{journal}{\emph{CoRR}}  \bibinfo{volume}{abs/1906.07155}
  (\bibinfo{year}{2019}).
\newblock


\bibitem[Cheng et~al\mbox{.}(2019)]%
        {DBLP:conf/mm/ChengLNP0Z19}
\bibfield{author}{\bibinfo{person}{Zhanzhan Cheng}, \bibinfo{person}{Jing Lu},
  \bibinfo{person}{Yi Niu}, \bibinfo{person}{Shiliang Pu}, \bibinfo{person}{Fei
  Wu}, {and} \bibinfo{person}{Shuigeng Zhou}.} \bibinfo{year}{2019}\natexlab{}.
\newblock \showarticletitle{You Only Recognize Once: Towards Fast Video Text
  Spotting}. In \bibinfo{booktitle}{\emph{ACM MM}}. \bibinfo{publisher}{{ACM}},
  \bibinfo{pages}{855--863}.
\newblock


\bibitem[Devlin et~al\mbox{.}(2019)]%
        {devlin2019bert}
\bibfield{author}{\bibinfo{person}{Jacob Devlin}, \bibinfo{person}{Ming{-}Wei
  Chang}, \bibinfo{person}{Kenton Lee}, {and} \bibinfo{person}{Kristina
  Toutanova}.} \bibinfo{year}{2019}\natexlab{}.
\newblock \showarticletitle{{BERT:} Pre-training of Deep Bidirectional
  Transformers for Language Understanding}. In
  \bibinfo{booktitle}{\emph{NAACL-HLT}}. \bibinfo{pages}{4171--4186}.
\newblock


\bibitem[Fang et~al\mbox{.}(2021)]%
        {DBLP:conf/cvpr/FangXWM021}
\bibfield{author}{\bibinfo{person}{Shancheng Fang}, \bibinfo{person}{Hongtao
  Xie}, \bibinfo{person}{Yuxin Wang}, \bibinfo{person}{Zhendong Mao}, {and}
  \bibinfo{person}{Yongdong Zhang}.} \bibinfo{year}{2021}\natexlab{}.
\newblock \showarticletitle{Read Like Humans: Autonomous, Bidirectional and
  Iterative Language Modeling for Scene Text Recognition}. In
  \bibinfo{booktitle}{\emph{CVPR}}. \bibinfo{pages}{7098--7107}.
\newblock


\bibitem[Guo et~al\mbox{.}(2009)]%
        {DBLP:conf/sigir/GuoXCL09}
\bibfield{author}{\bibinfo{person}{Jiafeng Guo}, \bibinfo{person}{Gu Xu},
  \bibinfo{person}{Xueqi Cheng}, {and} \bibinfo{person}{Hang Li}.}
  \bibinfo{year}{2009}\natexlab{}.
\newblock \showarticletitle{Named entity recognition in query}. In
  \bibinfo{booktitle}{\emph{SIGIR}}. \bibinfo{pages}{267--274}.
\newblock


\bibitem[He et~al\mbox{.}(2017)]%
        {he2017mask}
\bibfield{author}{\bibinfo{person}{Kaiming He}, \bibinfo{person}{Georgia
  Gkioxari}, \bibinfo{person}{Piotr Dollar}, {and} \bibinfo{person}{Ross
  Girshick}.} \bibinfo{year}{2017}\natexlab{}.
\newblock \showarticletitle{{Mask R-CNN}}. In \bibinfo{booktitle}{\emph{ICCV}}.
  \bibinfo{pages}{2980--2988}.
\newblock


\bibitem[Hochreiter and Schmidhuber(1997)]%
        {hochreiter1997long}
\bibfield{author}{\bibinfo{person}{Sepp Hochreiter} {and}
  \bibinfo{person}{J{\"{u}}rgen Schmidhuber}.} \bibinfo{year}{1997}\natexlab{}.
\newblock \showarticletitle{Long Short-Term Memory}.
\newblock \bibinfo{journal}{\emph{Neural Computation}} \bibinfo{volume}{9},
  \bibinfo{number}{8} (\bibinfo{year}{1997}), \bibinfo{pages}{1735--1780}.
\newblock


\bibitem[Huang et~al\mbox{.}(2015)]%
        {DBLP:journals/corr/HuangXY15}
\bibfield{author}{\bibinfo{person}{Zhiheng Huang}, \bibinfo{person}{Wei Xu},
  {and} \bibinfo{person}{Kai Yu}.} \bibinfo{year}{2015}\natexlab{}.
\newblock \showarticletitle{Bidirectional {LSTM-CRF} Models for Sequence
  Tagging}.
\newblock \bibinfo{journal}{\emph{CoRR}} (\bibinfo{year}{2015}).
\newblock


\bibitem[Jaderberg et~al\mbox{.}(2015)]%
        {DBLP:conf/nips/JaderbergSZK15}
\bibfield{author}{\bibinfo{person}{Max Jaderberg}, \bibinfo{person}{Karen
  Simonyan}, \bibinfo{person}{Andrew Zisserman}, {and} \bibinfo{person}{Koray
  Kavukcuoglu}.} \bibinfo{year}{2015}\natexlab{}.
\newblock \showarticletitle{Spatial Transformer Networks}. In
  \bibinfo{booktitle}{\emph{NeurIPS}}. \bibinfo{pages}{2017--2025}.
\newblock


\bibitem[Jauhar et~al\mbox{.}(2016)]%
        {DBLP:conf/acl/JauharTH16}
\bibfield{author}{\bibinfo{person}{Sujay~Kumar Jauhar},
  \bibinfo{person}{Peter~D. Turney}, {and} \bibinfo{person}{Eduard~H. Hovy}.}
  \bibinfo{year}{2016}\natexlab{}.
\newblock \showarticletitle{Tables as Semi-structured Knowledge for Question
  Answering}. In \bibinfo{booktitle}{\emph{ACL}}.
\newblock


\bibitem[Katti et~al\mbox{.}(2018)]%
        {DBLP:conf/emnlp/KattiRGBBHF18}
\bibfield{author}{\bibinfo{person}{Anoop~R. Katti}, \bibinfo{person}{Christian
  Reisswig}, \bibinfo{person}{Cordula Guder}, \bibinfo{person}{Sebastian
  Brarda}, \bibinfo{person}{Steffen Bickel}, \bibinfo{person}{Johannes
  H{\"{o}}hne}, {and} \bibinfo{person}{Jean~Baptiste Faddoul}.}
  \bibinfo{year}{2018}\natexlab{}.
\newblock \showarticletitle{Chargrid: Towards Understanding 2D Documents}. In
  \bibinfo{booktitle}{\emph{EMNLP}}. \bibinfo{pages}{4459--4469}.
\newblock


\bibitem[Khan et~al\mbox{.}(2019)]%
        {khan2019table}
\bibfield{author}{\bibinfo{person}{Saqib~Ali Khan}, \bibinfo{person}{Syed
  Muhammad~Daniyal Khalid}, \bibinfo{person}{Muhammad~Ali Shahzad}, {and}
  \bibinfo{person}{Faisal Shafait}.} \bibinfo{year}{2019}\natexlab{}.
\newblock \showarticletitle{Table Structure Extraction with Bi-Directional
  Gated Recurrent Unit Networks}. In \bibinfo{booktitle}{\emph{ICDAR}}.
  \bibinfo{pages}{1366--1371}.
\newblock


\bibitem[Li et~al\mbox{.}(2017)]%
        {li2017towards}
\bibfield{author}{\bibinfo{person}{Hui Li}, \bibinfo{person}{Peng Wang}, {and}
  \bibinfo{person}{Chunhua Shen}.} \bibinfo{year}{2017}\natexlab{}.
\newblock \showarticletitle{{Towards End-to-end Text Spotting with
  Convolutional Recurrent Neural Networks}}. In
  \bibinfo{booktitle}{\emph{ICCV}}. \bibinfo{pages}{5248--5256}.
\newblock


\bibitem[Li et~al\mbox{.}(2020)]%
        {DBLP:conf/eccv/LiGBWYZ20}
\bibfield{author}{\bibinfo{person}{Liangcheng Li}, \bibinfo{person}{Feiyu Gao},
  \bibinfo{person}{Jiajun Bu}, \bibinfo{person}{Yongpan Wang},
  \bibinfo{person}{Zhi Yu}, {and} \bibinfo{person}{Qi Zheng}.}
  \bibinfo{year}{2020}\natexlab{}.
\newblock \showarticletitle{An End-to-End {OCR} Text Re-organization Sequence
  Learning for Rich-Text Detail Image Comprehension}. In
  \bibinfo{booktitle}{\emph{ECCV}}, Vol.~\bibinfo{volume}{12370}.
  \bibinfo{pages}{85--100}.
\newblock


\bibitem[Liao et~al\mbox{.}(2019)]%
        {liao2019mask}
\bibfield{author}{\bibinfo{person}{Minghui Liao}, \bibinfo{person}{Pengyuan
  Lyu}, \bibinfo{person}{Minghang He}, \bibinfo{person}{Cong Yao},
  \bibinfo{person}{Wenhao Wu}, {and} \bibinfo{person}{Xiang Bai}.}
  \bibinfo{year}{2019}\natexlab{}.
\newblock \showarticletitle{Mask textspotter: An end-to-end trainable neural
  network for spotting text with arbitrary shapes}.
\newblock \bibinfo{journal}{\emph{IEEE TPAMI}} \bibinfo{volume}{1},
  \bibinfo{number}{1} (\bibinfo{year}{2019}).
\newblock


\bibitem[Liao et~al\mbox{.}(2020)]%
        {DBLP:conf/aaai/LiaoWYCB20}
\bibfield{author}{\bibinfo{person}{Minghui Liao}, \bibinfo{person}{Zhaoyi Wan},
  \bibinfo{person}{Cong Yao}, \bibinfo{person}{Kai Chen}, {and}
  \bibinfo{person}{Xiang Bai}.} \bibinfo{year}{2020}\natexlab{}.
\newblock \showarticletitle{Real-Time Scene Text Detection with Differentiable
  Binarization}. In \bibinfo{booktitle}{\emph{AAAI}}.
  \bibinfo{pages}{11474--11481}.
\newblock


\bibitem[Mathew et~al\mbox{.}(2021)]%
        {DBLP:conf/wacv/MathewKJ21}
\bibfield{author}{\bibinfo{person}{Minesh Mathew}, \bibinfo{person}{Dimosthenis
  Karatzas}, {and} \bibinfo{person}{C.~V. Jawahar}.}
  \bibinfo{year}{2021}\natexlab{}.
\newblock \showarticletitle{DocVQA: {A} Dataset for {VQA} on Document Images}.
  In \bibinfo{booktitle}{\emph{WACV}}. \bibinfo{pages}{2199--2208}.
\newblock


\bibitem[Qiao et~al\mbox{.}(2021a)]%
        {DBLP:conf/aaai/QiaoCCXNPW21}
\bibfield{author}{\bibinfo{person}{Liang Qiao}, \bibinfo{person}{Ying Chen},
  \bibinfo{person}{Zhanzhan Cheng}, \bibinfo{person}{Yunlu Xu},
  \bibinfo{person}{Yi Niu}, \bibinfo{person}{Shiliang Pu}, {and}
  \bibinfo{person}{Fei Wu}.} \bibinfo{year}{2021}\natexlab{a}.
\newblock \showarticletitle{{MANGO:} {A} Mask Attention Guided One-Stage Scene
  Text Spotter}. In \bibinfo{booktitle}{\emph{AAAI}}.
  \bibinfo{pages}{2467--2476}.
\newblock


\bibitem[Qiao et~al\mbox{.}(2021b)]%
        {DBLP:conf/icdar/QiaoLCZPNRT021}
\bibfield{author}{\bibinfo{person}{Liang Qiao}, \bibinfo{person}{Zaisheng Li},
  \bibinfo{person}{Zhanzhan Cheng}, \bibinfo{person}{Peng Zhang},
  \bibinfo{person}{Shiliang Pu}, \bibinfo{person}{Yi Niu},
  \bibinfo{person}{Wenqi Ren}, \bibinfo{person}{Wenming Tan}, {and}
  \bibinfo{person}{Fei Wu}.} \bibinfo{year}{2021}\natexlab{b}.
\newblock \showarticletitle{{LGPMA:} Complicated Table Structure Recognition
  with Local and Global Pyramid Mask Alignment}. In
  \bibinfo{booktitle}{\emph{ICDAR}}, Vol.~\bibinfo{volume}{12821}.
  \bibinfo{pages}{99--114}.
\newblock


\bibitem[Qiao et~al\mbox{.}(2020)]%
        {DBLP:conf/aaai/QiaoTCXNPW20}
\bibfield{author}{\bibinfo{person}{Liang Qiao}, \bibinfo{person}{Sanli Tang},
  \bibinfo{person}{Zhanzhan Cheng}, \bibinfo{person}{Yunlu Xu},
  \bibinfo{person}{Yi Niu}, \bibinfo{person}{Shiliang Pu}, {and}
  \bibinfo{person}{Fei Wu}.} \bibinfo{year}{2020}\natexlab{}.
\newblock \showarticletitle{Text Perceptron: Towards End-to-End
  Arbitrary-Shaped Text Spotting}. In \bibinfo{booktitle}{\emph{AAAI}}.
  \bibinfo{pages}{11899--11907}.
\newblock


\bibitem[Redmon and Farhadi(2017)]%
        {DBLP:conf/cvpr/RedmonF17}
\bibfield{author}{\bibinfo{person}{Joseph Redmon} {and} \bibinfo{person}{Ali
  Farhadi}.} \bibinfo{year}{2017}\natexlab{}.
\newblock \showarticletitle{{YOLO9000:} Better, Faster, Stronger}. In
  \bibinfo{booktitle}{\emph{CVPR}}. \bibinfo{publisher}{{IEEE} Computer
  Society}, \bibinfo{pages}{6517--6525}.
\newblock


\bibitem[Sang and Meulder(2003)]%
        {DBLP:conf/conll/SangM03}
\bibfield{author}{\bibinfo{person}{Erik F. Tjong~Kim Sang} {and}
  \bibinfo{person}{Fien~De Meulder}.} \bibinfo{year}{2003}\natexlab{}.
\newblock \showarticletitle{Introduction to the CoNLL-2003 Shared Task:
  Language-Independent Named Entity Recognition}. In
  \bibinfo{booktitle}{\emph{CoNLL/HLT-NAACL}}. \bibinfo{pages}{142--147}.
\newblock


\bibitem[Shi et~al\mbox{.}(2017)]%
        {CRNN}
\bibfield{author}{\bibinfo{person}{Baoguang Shi}, \bibinfo{person}{Xiang Bai},
  {and} \bibinfo{person}{Cong Yao}.} \bibinfo{year}{2017}\natexlab{}.
\newblock \showarticletitle{An End-to-End Trainable Neural Network for
  Image-Based Sequence Recognition and Its Application to Scene Text
  Recognition}.
\newblock \bibinfo{journal}{\emph{IEEE TPAMI}} \bibinfo{volume}{39},
  \bibinfo{number}{11} (\bibinfo{year}{2017}), \bibinfo{pages}{2298--2304}.
\newblock


\bibitem[Shi et~al\mbox{.}(2016)]%
        {DBLP:conf/cvpr/ShiWLYB16}
\bibfield{author}{\bibinfo{person}{Baoguang Shi}, \bibinfo{person}{Xinggang
  Wang}, \bibinfo{person}{Pengyuan Lyu}, \bibinfo{person}{Cong Yao}, {and}
  \bibinfo{person}{Xiang Bai}.} \bibinfo{year}{2016}\natexlab{}.
\newblock \showarticletitle{Robust Scene Text Recognition with Automatic
  Rectification}. In \bibinfo{booktitle}{\emph{CVPR}}.
  \bibinfo{pages}{4168--4176}.
\newblock


\bibitem[Sun et~al\mbox{.}(2021)]%
        {DBLP:journals/corr/abs-2103-14470}
\bibfield{author}{\bibinfo{person}{Hongbin Sun}, \bibinfo{person}{Zhanghui
  Kuang}, \bibinfo{person}{Xiaoyu Yue}, \bibinfo{person}{Chenhao Lin}, {and}
  \bibinfo{person}{Wayne Zhang}.} \bibinfo{year}{2021}\natexlab{}.
\newblock \showarticletitle{Spatial Dual-Modality Graph Reasoning for Key
  Information Extraction}.
\newblock \bibinfo{journal}{\emph{CoRR}}  \bibinfo{volume}{abs/2103.14470}
  (\bibinfo{year}{2021}).
\newblock


\bibitem[Wang et~al\mbox{.}(2021b)]%
        {DBLP:conf/aaai/WangZQLZLHLDS21}
\bibfield{author}{\bibinfo{person}{Pengfei Wang}, \bibinfo{person}{Chengquan
  Zhang}, \bibinfo{person}{Fei Qi}, \bibinfo{person}{Shanshan Liu},
  \bibinfo{person}{Xiaoqiang Zhang}, \bibinfo{person}{Pengyuan Lyu},
  \bibinfo{person}{Junyu Han}, \bibinfo{person}{Jingtuo Liu},
  \bibinfo{person}{Errui Ding}, {and} \bibinfo{person}{Guangming Shi}.}
  \bibinfo{year}{2021}\natexlab{b}.
\newblock \showarticletitle{PGNet: Real-time Arbitrarily-Shaped Text Spotting
  with Point Gathering Network}. In \bibinfo{booktitle}{\emph{AAAI}}.
  \bibinfo{pages}{2782--2790}.
\newblock


\bibitem[Wang et~al\mbox{.}(2020)]%
        {wang2020structure}
\bibfield{author}{\bibinfo{person}{Zhiruo Wang}, \bibinfo{person}{Haoyu Dong},
  \bibinfo{person}{Ran Jia}, \bibinfo{person}{Jia Li}, \bibinfo{person}{Zhiyi
  Fu}, \bibinfo{person}{Shi Han}, {and} \bibinfo{person}{Dongmei Zhang}.}
  \bibinfo{year}{2020}\natexlab{}.
\newblock \showarticletitle{Structure-aware Pre-training for Table
  Understanding with Tree-based Transformers}.
\newblock \bibinfo{journal}{\emph{CoRR}}  \bibinfo{volume}{abs/2010.12537}
  (\bibinfo{year}{2020}).
\newblock


\bibitem[Wang et~al\mbox{.}(2021a)]%
        {wang2021layoutreader}
\bibfield{author}{\bibinfo{person}{Zilong Wang}, \bibinfo{person}{Yiheng Xu},
  \bibinfo{person}{Lei Cui}, \bibinfo{person}{Jingbo Shang}, {and}
  \bibinfo{person}{Furu Wei}.} \bibinfo{year}{2021}\natexlab{a}.
\newblock \showarticletitle{LayoutReader: Pre-training of Text and Layout for
  Reading Order Detection}. In \bibinfo{booktitle}{\emph{EMNLP}}.
  \bibinfo{pages}{4735--4744}.
\newblock


\bibitem[Yan et~al\mbox{.}(2021)]%
        {DBLP:conf/acl/YanGDGZQ20}
\bibfield{author}{\bibinfo{person}{Hang Yan}, \bibinfo{person}{Tao Gui},
  \bibinfo{person}{Junqi Dai}, \bibinfo{person}{Qipeng Guo},
  \bibinfo{person}{Zheng Zhang}, {and} \bibinfo{person}{Xipeng Qiu}.}
  \bibinfo{year}{2021}\natexlab{}.
\newblock \showarticletitle{A Unified Generative Framework for Various {NER}
  Subtasks}. In \bibinfo{booktitle}{\emph{ACL}}. \bibinfo{pages}{5808--5822}.
\newblock


\bibitem[Yang et~al\mbox{.}(2017)]%
        {DBLP:conf/cvpr/YangYAKKG17}
\bibfield{author}{\bibinfo{person}{Xiao Yang}, \bibinfo{person}{Ersin Yumer},
  \bibinfo{person}{Paul Asente}, \bibinfo{person}{Mike Kraley},
  \bibinfo{person}{Daniel Kifer}, {and} \bibinfo{person}{C.~Lee Giles}.}
  \bibinfo{year}{2017}\natexlab{}.
\newblock \showarticletitle{Learning to Extract Semantic Structure from
  Documents Using Multimodal Fully Convolutional Neural Networks}. In
  \bibinfo{booktitle}{\emph{CVPR}}. \bibinfo{pages}{4342--4351}.
\newblock


\bibitem[Zhang et~al\mbox{.}(2021b)]%
        {DBLP:conf/aaai/0003XCPNWZ21}
\bibfield{author}{\bibinfo{person}{Chengwei Zhang}, \bibinfo{person}{Yunlu Xu},
  \bibinfo{person}{Zhanzhan Cheng}, \bibinfo{person}{Shiliang Pu},
  \bibinfo{person}{Yi Niu}, \bibinfo{person}{Fei Wu}, {and}
  \bibinfo{person}{Futai Zou}.} \bibinfo{year}{2021}\natexlab{b}.
\newblock \showarticletitle{{SPIN:} Structure-Preserving Inner Offset Network
  for Scene Text Recognition}. In \bibinfo{booktitle}{\emph{AAAI}}.
  \bibinfo{publisher}{{AAAI} Press}, \bibinfo{pages}{3305--3314}.
\newblock


\bibitem[Zhang et~al\mbox{.}(2021a)]%
        {DBLP:conf/icdar/ZhangLQCPN021}
\bibfield{author}{\bibinfo{person}{Peng Zhang}, \bibinfo{person}{Can Li},
  \bibinfo{person}{Liang Qiao}, \bibinfo{person}{Zhanzhan Cheng},
  \bibinfo{person}{Shiliang Pu}, \bibinfo{person}{Yi Niu}, {and}
  \bibinfo{person}{Fei Wu}.} \bibinfo{year}{2021}\natexlab{a}.
\newblock \showarticletitle{{VSR:} {A} Unified Framework for Document Layout
  Analysis Combining Vision, Semantics and Relations}. In
  \bibinfo{booktitle}{\emph{ICDAR}}, Vol.~\bibinfo{volume}{12821}.
  \bibinfo{pages}{115--130}.
\newblock


\bibitem[Zhang et~al\mbox{.}(2020)]%
        {DBLP:conf/mm/ZhangXCP0QNW20}
\bibfield{author}{\bibinfo{person}{Peng Zhang}, \bibinfo{person}{Yunlu Xu},
  \bibinfo{person}{Zhanzhan Cheng}, \bibinfo{person}{Shiliang Pu},
  \bibinfo{person}{Jing Lu}, \bibinfo{person}{Liang Qiao}, \bibinfo{person}{Yi
  Niu}, {and} \bibinfo{person}{Fei Wu}.} \bibinfo{year}{2020}\natexlab{}.
\newblock \showarticletitle{{TRIE:} End-to-End Text Reading and Information
  Extraction for Document Understanding}. In \bibinfo{booktitle}{\emph{ACM
  MM}}. \bibinfo{publisher}{{ACM}}, \bibinfo{pages}{1413--1422}.
\newblock


\bibitem[Zhong et~al\mbox{.}(2019)]%
        {DBLP:conf/icdar/ZhongTJ19}
\bibfield{author}{\bibinfo{person}{Xu Zhong}, \bibinfo{person}{Jianbin Tang},
  {and} \bibinfo{person}{Antonio Jimeno{-}Yepes}.}
  \bibinfo{year}{2019}\natexlab{}.
\newblock \showarticletitle{PubLayNet: Largest Dataset Ever for Document Layout
  Analysis}. In \bibinfo{booktitle}{\emph{ICDAR}}. \bibinfo{publisher}{{IEEE}},
  \bibinfo{pages}{1015--1022}.
\newblock


\bibitem[Zhou et~al\mbox{.}(2017)]%
        {zhou2017east}
\bibfield{author}{\bibinfo{person}{Xinyu Zhou}, \bibinfo{person}{Cong Yao},
  \bibinfo{person}{He Wen}, \bibinfo{person}{Yuzhi Wang},
  \bibinfo{person}{Shuchang Zhou}, \bibinfo{person}{Weiran He}, {and}
  \bibinfo{person}{Jiajun Liang}.} \bibinfo{year}{2017}\natexlab{}.
\newblock \showarticletitle{{EAST: An Efficient and Accurate Scene Text
  Detector}}. In \bibinfo{booktitle}{\emph{CVPR}}. \bibinfo{pages}{2642--2651}.
\newblock


\end{thebibliography}

\end{document}